\documentclass[conference]{IEEEtran}
\IEEEoverridecommandlockouts
\newcommand{\shortname}{\texttt{WaSA}}
\usepackage{booktabs, adjustbox, multirow, rotating}
\newcommand{\best}[1]{\textbf{#1}}
\newcommand{\ora}[1]{\underline{#1}}
\usepackage{subcaption}
\usepackage{cite}

\usepackage{amsmath,amssymb,amsfonts}
\usepackage{algorithmicx}
\usepackage{graphicx}
\usepackage{textcomp}
\usepackage{xcolor}
\usepackage{algorithm} 
\usepackage{url}   
\usepackage{algpseudocode}  
\def\BibTeX{{\rm B\kern-.05em{\sc i\kern-.025em b}\kern-.08em
    T\kern-.1667em\lower.7ex\hbox{E}\kern-.125emX}}

\let\oldthebibliography\thebibliography
\renewcommand{\thebibliography}[1]{%
  \oldthebibliography{#1}%
  \setlength{\itemsep}{0pt}\setlength{\parskip}{0pt}}

\begin{document}
\title{\shortname{}: Wavelet Scattering Autoencoders for Unsupervised Offline Detection of Malicious Clients in Federated Learning}
\author{\IEEEauthorblockN{Alessandro Licciardi}
\IEEEauthorblockA{\textit{Department of Mathematical Sciences, Politecnico di Torino, Turin, IT} \\
\textit{National Institute of Nuclear Physics (INFN), Sezione di Torino, Turin, IT}\\
\texttt{alessandro.licciardi@polito.it}
\thanks{979-8-3195-0489-0/26/\$31.00 ©2026 IEEE}}
 }
\maketitle
\vspace{-1em}
\begin{abstract}
Federated Learning (FL) enables collaborative model training without exposing raw data, yet remains vulnerable to malicious clients injecting poisoned datasets. Existing defenses are typically reactive (operating during training) or supervised (requiring prior knowledge of attack types)---assumptions that rarely hold in open-ended IoT deployments. We propose \shortname{}, a fully unsupervised, offline framework for malicious client detection. Before training, each client transmits a single, privacy-preserving spectral summary of its data to the server. A denoising autoencoder, trained solely on clean auxiliary data, scores clients based on their deviation from the learned benign manifold, identifying corruptions regardless of the attack type. \shortname{} requires no attack labels, generalizes to unseen anomalies, and introduces a Boltzmann sampling strategy for client selection that smoothly interpolates between hard exclusion and standard aggregation. Experiments on MNIST, CIFAR-10, and CIFAR-100 under diverse attack scenarios (up to 90\% malicious clients) demonstrate consistent improvements over supervised baselines and robust aggregation methods, with negligible communication overhead. Code is available at \url{https://github.com/alelicciardi99/wavelet_autoencoders_malicious}.
\end{abstract}
\vspace{-1.5em}

\section{Introduction}
Federated Learning (FL) enables decentralized model training without exposing raw data~\cite{mcmahan2017communication, bonawitz2019towards}. 
This privacy guarantee, however, introduces a structural vulnerability: malicious or faulty clients can inject poisoned local datasets, silently degrading global model performance~\cite{li2020federated, yin2018byzantine, cao2021fltrust}. 
Proactively detecting these clients before training begins is critical, but remains an open challenge when attack signatures are unknown.

Most defenses operate \emph{during} active training, weaving anomaly detection into the synchronous aggregation loop---a coupling that introduces computational bottlenecks and leaves the global model exposed to contamination before any attacker can be isolated~\cite{guerraoui2018hidden}. 
An offline alternative bypasses both issues: screening clients \emph{before} the first communication round separates defensive computation from optimization entirely. 
Introducing auxiliary data to establish expected baselines~\cite{sattler2021fedaux, goswamicovariances} allows the server to filter malicious participants at ``round zero,'' ensuring training commences with a sanitized client pool.

While one-shot offline detection outperforms reactive strategies, existing methods often rely on \emph{supervised} classifiers trained with simulated attack labels~\cite{licciardi2025waffle}. 
The implicit requirement that the server knows the attack type in advance, however, rarely holds. Real-world corruptions such as sensor degradation or novel adversarial perturbations emerge without warning, meaning a detector tied to specific attack signatures must be constantly retrained whenever the threat model shifts.

\textbf{This work.} 
We present \shortname{} (\emph{Wavelet Scattering Autoencoder}), a fully unsupervised, attack-agnostic framework. 
Rather than relying on labeled attacks, \shortname{} trains a denoising autoencoder exclusively on \emph{clean} spectral prototypes derived from an auxiliary dataset. 
Since this step precedes federated training entirely, it incurs no overhead on the active FL rounds. 
At detection time, clients whose Wavelet Scattering Transform (WST) embeddings deviate from the learned benign manifold receive a high normalized reconstruction error $\mathcal{E}_k$ and are flagged. 
To ensure robustness, \shortname{} avoids theoretically unstable spectral components (e.g., Fourier branches~\cite{mallat2012group}) in favor of pure WST, and introduces a Boltzmann sampling strategy for probabilistic client participation.

\textbf{Related Work.} 
Robust aggregation methods (e.g., Krum~\cite{blanchard2017machine}, TrimmedMean~\cite{yin2018byzantine}, FLTrust~\cite{cao2021fltrust}) mitigate malicious updates without explicit identification, but fail theoretically when attackers exceed 50\%~\cite{blanchard2017machine}. 
Online detection methods~\cite{zhang2022fldetector, wang2024revisiting} identify attackers via multi-round update consistency. 
This reactive paradigm demands sustained communication overhead---a significant burden in IoT deployments---and leaves the model transiently contaminated. 
Offline detection circumvents these limits; in particular, \textsc{Waffle}~\cite{licciardi2025waffle} filters clients before the FL training begins, using spectral embeddings computed with WST and Fourier Transform but requires a supervised classifier. \shortname{} extends this offline direction to a fully unsupervised setting.

\textbf{Contributions.}
\begin{itemize}
    \item We propose \shortname{}, the first \emph{unsupervised} offline detector for FL. A denoising autoencoder trained on clean WST embeddings detects anomalous clients at round zero via a normalized reconstruction error $\mathcal{E}_k$, requiring no prior attack knowledge.
    \item We introduce a GPU-accelerated, SVD-based spectral prototype that efficiently compresses a client's data distribution into a single privacy-preserving image.
    \item We propose a statistical mechanics-inspired \emph{Boltzmann sampling} strategy for FL client selection, interpolating between hard exclusion and uniform sampling via a temperature parameter ${\mathcal{T}}$.
\end{itemize}
\vspace{-1em}
\section{Methodology}
\label{sec:methodology}
In standard FL, a central server coordinates $K$ clients to minimize a global objective function $L(\theta) = \sum_{k=1}^K p_k L_k(\theta)$, where $p_k = n_k/n$ is the relative data weight of client $k$ ($n = \sum_k n_k$), and $L_k(\theta)$ is its local empirical risk. Under the standard \texttt{FedAvg} algorithm~\cite{mcmahan2017communication}, at each communication round $t$, a selected subset of clients performs local optimization on the global model $\theta_t$ to produce updated weights $\theta_{t+1}^{(k)}$. The server subsequently aggregates the updates, i.e. , $\theta_{t+1} = \sum_{k} p_k \theta_{t+1}^{(k)}$.
\shortname{} operates over two distinct phases. In the offline phase, the server approximates the compact manifold of benign client data by minimizing a denoising empirical risk objective over synthetic WST embeddings derived from an auxiliary dataset. Before the federated training starts, clients perform a locally optimal rank-reduced projection of their data, transforming it via wavelet scattering operator. The server subsequently evaluates these representations by computing their reconstruction error, which formally quantifies their structural deviation from the expected benign distribution. These scalar scores are then utilized strictly to govern federated participation through either deterministic thresholding or probabilistic Boltzmann sampling.

\subsection{Client-Side Optimal Rank-Reduced Representation}
\label{subsec:client_representation}

Consider a federation of $K$ clients. Each client $k \in \{1, \ldots, K\}$ has its own local dataset $\{x_i^{(k)}\}_{i=1}^{n_k}$, where $x_i^{(k)} \in \mathbb{R}^{C \times H \times W}$. To generate a privacy-preserving and communication-efficient summary of this local distribution,  \cite{goswamicovariances}, the client computes a single spectral prototype $\hat{x}_k$, defined as the variance-weighted superposition of the empirical principal directions.

Each local dataset is flatten to form the matrix $X^{(k)} \in \mathbb{R}^{n_k \times D}$, where $D = C \cdot H \cdot W$, and define the centered matrix $\tilde{X}^{(k)} = X^{(k)} - \mathbf{1}_{n_k}{\mu_k}^\top$, where $\mu_k = \frac{1}{n_k}\sum_{i=1}^{n_k} x_i^{(k)}$ is the sample mean, and $\mathbf{1}_{n_k}$ a column vector of unitary entries. According to the Eckart-Young-Mirsky theorem~\cite{eckart1936approximation}, the rank-$r$ matrix that optimally approximates $\tilde{X}^{(k)}$ in the Frobenius norm is given by the truncated SVD ~\cite{golub1965calculating}:
\(
    \tilde{X}^{(k)}_r = U^{(k)}_r \Sigma^{(k)}_r {V^{(k)}_r}^\top
\)
where $V^{(k)}_r \in \mathbb{R}^{D \times r}$ contains the top-$r$ right singular vectors $\{v_i^{(k)}\}_{i=1}^r$, and $\Sigma^{(k)}_r$ contains the corresponding singular values $\sigma_1^{(k)} \geq \cdots \geq \sigma_r^{(k)} > 0$. The spectral prototype is subsequently constructed as:
\begin{equation}
    \hat{x}_k = \mu_k + \sum_{i=1}^r \alpha_i^{(k)}\, v_i^{(k)}, \qquad \alpha_i^{(k)} = \frac{{\sigma_i^{(k)}}^2}{\sum_{j=1}^r {\sigma_j^{(k)}}^2}
    \label{eq:prototype}
\end{equation}
where $\alpha_i^{(k)}$ quantifies the fraction of total variance captured by the $i$-th principal component, and $v_i^{(k)}$ is reshaped back to the original space $\mathbb{R}^{C \times H \times W}$. 

\subsection{Wavelet Scattering Transform}
\label{subsec:wst}

While the prototype $\hat{x}_k$ provides a dense linear summary of the client's data, it remains susceptible to local geometric deformations (e.g., small translations or rotations) that artificially inflate distances in the pixel space. To construct a representation strictly stable to such deformations, we project $\hat{x}_k$ through a Wavelet Scattering Transform~\cite{mallat2012group, bruna2013invariant}.

Let $\psi$ be a mother wavelet. We define a family of dilated and rotated wavelets $\psi_\lambda(u) = 2^{2j} \psi(2^j r^{-1} u)$, where the multi-index $\lambda = (j, r)$ incorporates the scale $2^j$ and rotation matrix $r$. For a maximal scale $2^J$, the scattering propagator along a path $p = (\lambda_1, \ldots, \lambda_m)$ is defined recursively for any signal $x \in L^2(\mathbb{R}^2)$ as:
\begin{equation}
    U[p]\,x = \bigl|\,U[(\lambda_1,\ldots,\lambda_{m-1})]\,x \star \psi_{\lambda_m}\bigr|, \quad U[\emptyset]\,x = x
\end{equation}
The scattering coefficients are obtained by applying a low-pass filter $\phi_J$ at scale $2^J$:
\(begin{equation})
    S^J[p]\,x = U[p]\,x \star \phi_J
\)
As rigorously proved in~\cite{mallat2012group} and expanded upon for FL anomaly detection in~\textsc{Waffle}~\cite{licciardi2025waffle}, the operator $S^J$ separates structurally distinct signals while maintaining Lipschitz stability to $C^2$-diffeomorphisms. For \shortname{}, we retain only the first-order scattering coefficients ($|\lambda| = 1$), vectorizing them to form the final client embedding:
\(
    z_k = \mathrm{vec}\!\left(\bigl\{S^J[p]\,\hat{x}_k : p \in \Lambda_J^{(1)}\bigr\}\right) \in \mathbb{R}^d.
    \label{eq:embedding}
\)
The non-invertibility of $S^J$, combined with the SVD truncation, inherently protects the raw local data from inversion attacks.

\subsection{Offline Manifold Learning via Denoising Autoencoder}
\label{subsec:dae}

To overcome the reliance on labeled attack profiles that restricts prior supervised defenses to known threats, \shortname{} employs a fully unsupervised denoising autoencoder~\cite{vincent2010stacked} to learn the distribution of benign clients. Let $f_\phi = \mathrm{Dec}_\phi \circ \mathrm{Enc}_\phi$ denote a neural network parameterized by $\phi$. The encoder $\mathrm{Enc}_\phi : \mathbb{R}^d \to \mathbb{R}^h$ compresses the WST embedding into a strict bottleneck ($h \ll d$), and the decoder $\mathrm{Dec}_\phi : \mathbb{R}^h \to \mathbb{R}^d$ reconstructs the original signal. By enforcing this strict bottleneck and training exclusively on clean data, the network is compelled to learn only the intrinsic low-dimensional manifold of benign representations. Consequently, anomalous embeddings that deviate from this manifold cannot be accurately reconstructed from the latent space, naturally exposing malicious clients without requiring prior attack knowledge.

Training is conducted exclusively offline using synthetic embeddings generated from an auxiliary dataset $D_{\text{aux}}$, as detailed in Algorithm~\ref{alg:wasa_training}. The parameters $\phi$ are optimized to minimize the expected empirical reconstruction error over the noise-corrupted batch $\tilde{Z}$:
\begin{equation}
    \mathcal{L}(\phi) = \frac{1}{B \cdot d} \left\|f_\phi\!\left(\tilde{Z}\right) - Z\right\|_F^2
    \label{eq:loss}
\end{equation}
where $\|\cdot\|_F$ denotes the Frobenious norm \cite{strang2006linear}.
Upon convergence to optimal weights $\phi^*$, the autoencoder defines a precise geometric boundary for expected benign behavior. We denote with $L_{\text{final}} = \mathcal{L}(\phi^*)$, the empirical loss at convergence, that will be considered as the baseline reconstruction error on clean data.

\begin{algorithm}[t]
\caption{\shortname{} Offline Training}
\label{alg:wasa_training}
\begin{algorithmic}[1]
\Require Auxiliary dataset $D_{\text{aux}}$, epochs $E$, batch size $B$, samples per client $n_s$, PCA rank $r$, WST operator $S^J$, noise $\sigma_{\text{noise}}$
\Ensure Parameters $\phi^{(E)}$, baseline loss $L_{\text{final}}$
\State Initialize $\phi^{(1)}$
\For{$e = 1, \dots, E$}
    \Statex\hspace{1.2em}\textit{// Phase 1: Synthetic Client Feature Extraction}
    \For{$b = 1, \dots, B$}
        \State Sample $\{x_i^{(b)}\}_{i=1}^{n_s} \sim D_{\text{aux}}$; compute mean $\mu^{(b)}$
        \State Center data $\tilde{X}^{(b)}$ and compute $\mathrm{SVD}_r(\tilde{X}^{(b)})$
        \State Form prototype $\hat{x}^{(b,e)}$ via Eq.~\eqref{eq:prototype}
        \State Compute WST embedding $z^{(b,e)} \in \mathbb{R}^d$ via Eq.~\eqref{eq:embedding}
    \EndFor
    \State Form clean batch $Z^{(e)} \leftarrow [z^{(1,e)}, \dots, z^{(B,e)}]^\top$
    \Statex\hspace{1.2em}\textit{// Phase 2: Denoising Autoencoder Update}
    \State Inject noise $\tilde{Z}^{(e)} \leftarrow Z^{(e)} + \mathcal{N}(0,\sigma_{\text{noise}}^2 I)$
    \State Compute loss $\mathcal{L}^{(e)} \leftarrow \frac{1}{Bd}\|f_{\phi^{(e)}}(\tilde{Z}^{(e)}) - Z^{(e)}\|_F^2$
    \State $\phi^{(e+1)} \leftarrow \texttt{Optimizer}(\phi^{(e)}, \nabla_{\phi}\mathcal{L}^{(e)})$
\EndFor
\State \Return $\phi^{(E)}$, $L_{\text{final}} \leftarrow \mathcal{L}^{(E)}$
\end{algorithmic}
\end{algorithm}

\subsection{Online Client Evaluation \& Filtering}
\label{subsec:online_filtering}

Prior to the first FL communication round, each client $k$ transmits its embedding $z_k$ to the server. The server computes the scale-free normalized anomaly score:
\begin{equation}
    \mathcal{E}_k = \frac{\left\|z_k - f_\phi(z_k)\right\|_2^2}{d \cdot L_{\text{final}}}
    \label{eq:score}
\end{equation}
Embeddings that map near the learned benign manifold yield $\mathcal{E}_k \lesssim 1$. Malicious clients exhibiting anomalous spectral structures yield $\mathcal{E}_k \gg 1$. \shortname{} utilizes these scores to govern participation via two variants:

\textbf{Deterministic Thresholding:} The server applies a hard threshold $\tau > 0$, defining the verified participating set as $\hat{\mathcal{B}} = \{k \in [K] : \mathcal{E}_k \leq \tau\}$. Aggregation is performed exclusively over $\hat{\mathcal{B}}$ for all subsequent FL rounds. A natural hard choice is $\tau = 1$.

\textbf{Boltzmann Sampling:} When deterministic thresholding is too rigid for highly heterogeneous deployments, we interpret $\mathcal{E}_k$ as an energy state and map it to a Gibbs distribution \cite{gallavotti2013statistical}, assigning a selection probability to each client:
\begin{equation}
    \pi_k = \frac{\exp(-\mathcal{E}_k / \mathcal{T})}{\sum_{j=1}^K \exp(-\mathcal{E}_j / \mathcal{T})}, \quad \mathcal{T} > 0
    \label{eq:boltzmann}
\end{equation}
where $T$ is the temperature hyperparameter -- smaller the value of $\mathcal{E}_k$ better the quality of the reconstruction . The server evaluates the full partition function over all $K$ clients prior to the first round. At each communication round $t$, a subset of clients is drawn without replacement according to $\pi_k$. This continuous interpolation allows the server to smoothly transition from hard exclusion ($\mathcal{T} \to 0$) to uniform random sampling ($\mathcal{T} \to \infty$).
Although \shortname{}-Boltzmann is primarily designed to bias client participation rather than to produce binary detection decisions, we adapt it for detection evaluation as follows. A client $k$ is classified as malicious if its expected participation is negligible over the entire training horizon, that is, if
\begin{equation}
    \pi_k
    < \frac{1}{P} \cdot \frac{T_{\max}}{T_{\text{total}}},
    \label{eq:boltzmann_threshold}
\end{equation}
where $P$ is the number of clients selected per round, $T_{\text{total}}$ is the total number of communication rounds, and $T_{\max}$ is the maximum number of rounds deemed harmless for a corrupted client. 

\vspace{-1em}
\section{Experiments and Results}

\begin{figure*}[t]
    \centering
    \includegraphics[width=\columnwidth]{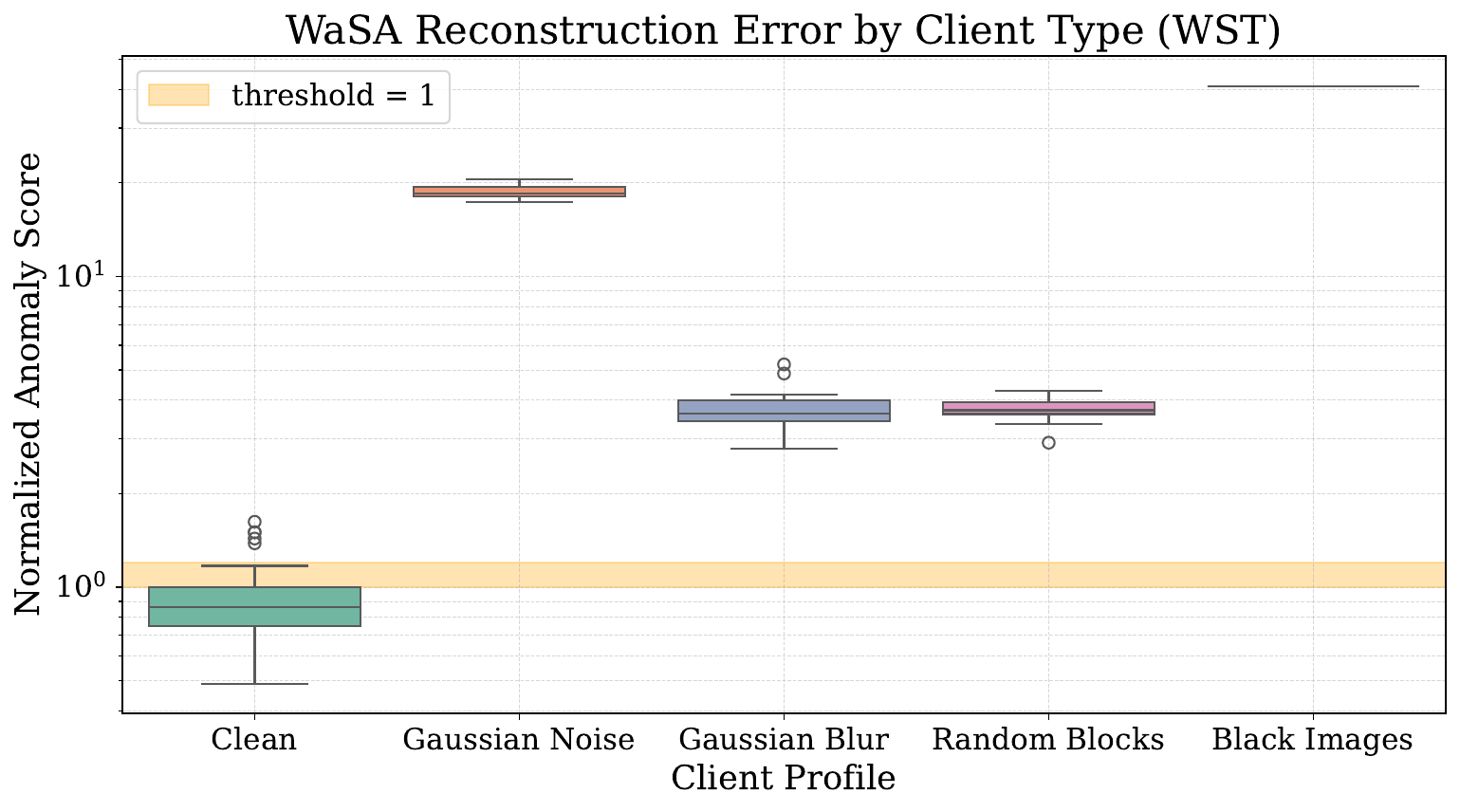}
    \hfill
    \includegraphics[width=\columnwidth]{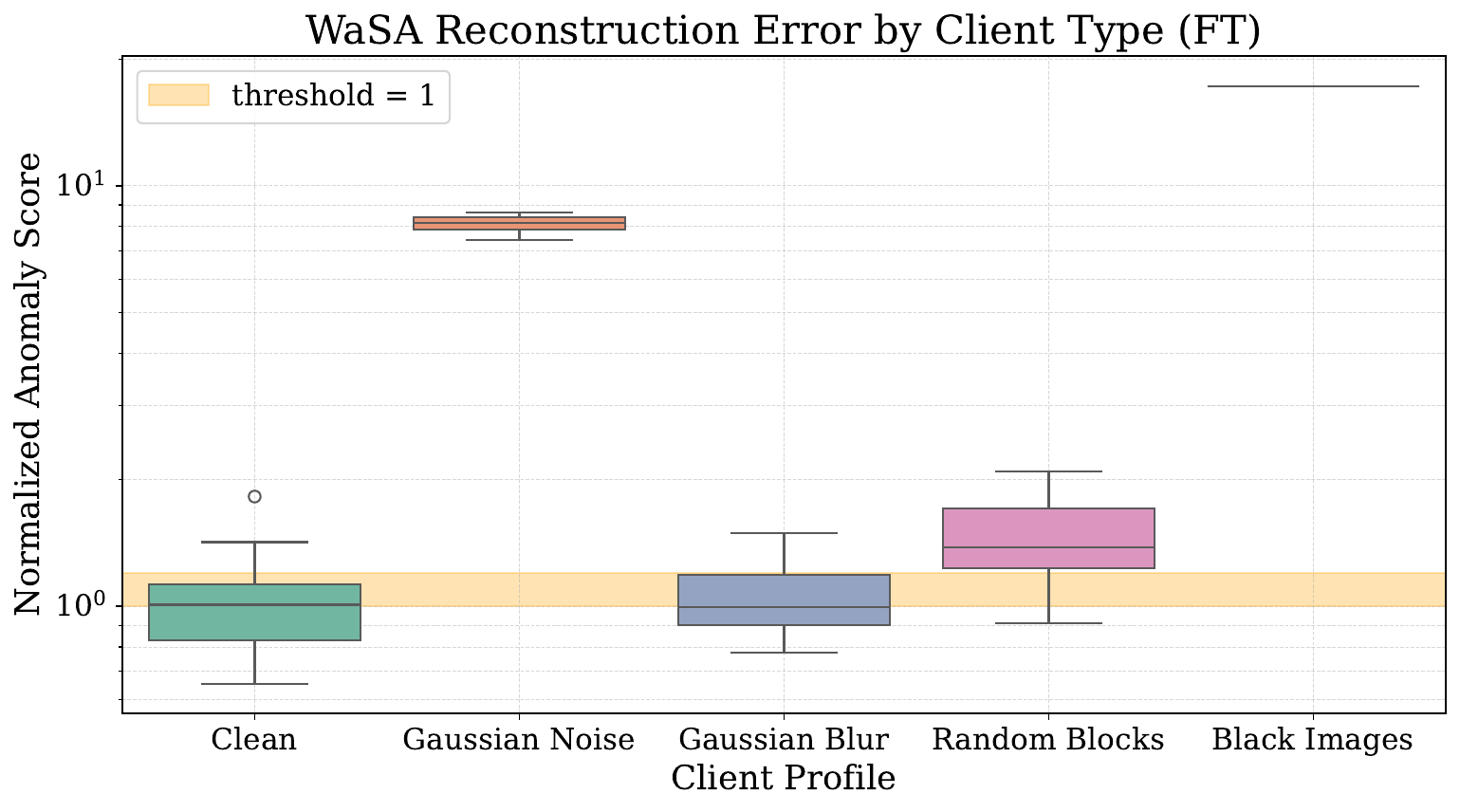}
    
    \caption{\small Normalized anomaly scores $\mathcal{E}_k$ for all 100 clients on CIFAR-10
under the 60\% corruption scenario. \textbf{Left:} \shortname{} with WST backbone.
\textbf{Right:} \shortname{} with FT backbone, instead of WST. The orange band marks the detection
threshold $\tau = 1$. The WST representation yields a clear spectral separation
between benign clients and all four corruption types, including structurally subtle
attacks such as Gaussian blur and random pixel masking. The FT backbone fails to produce
a consistent anomaly signal, with corruption scores overlapping significantly with the
benign distribution, confirming the theoretical instability of the Fourier representation
to local deformations~\cite{mallat2012group}.}
    \label{fig:wstvsft}
\end{figure*}
\begin{figure*}
    \centering
    \includegraphics[width=.85\linewidth]{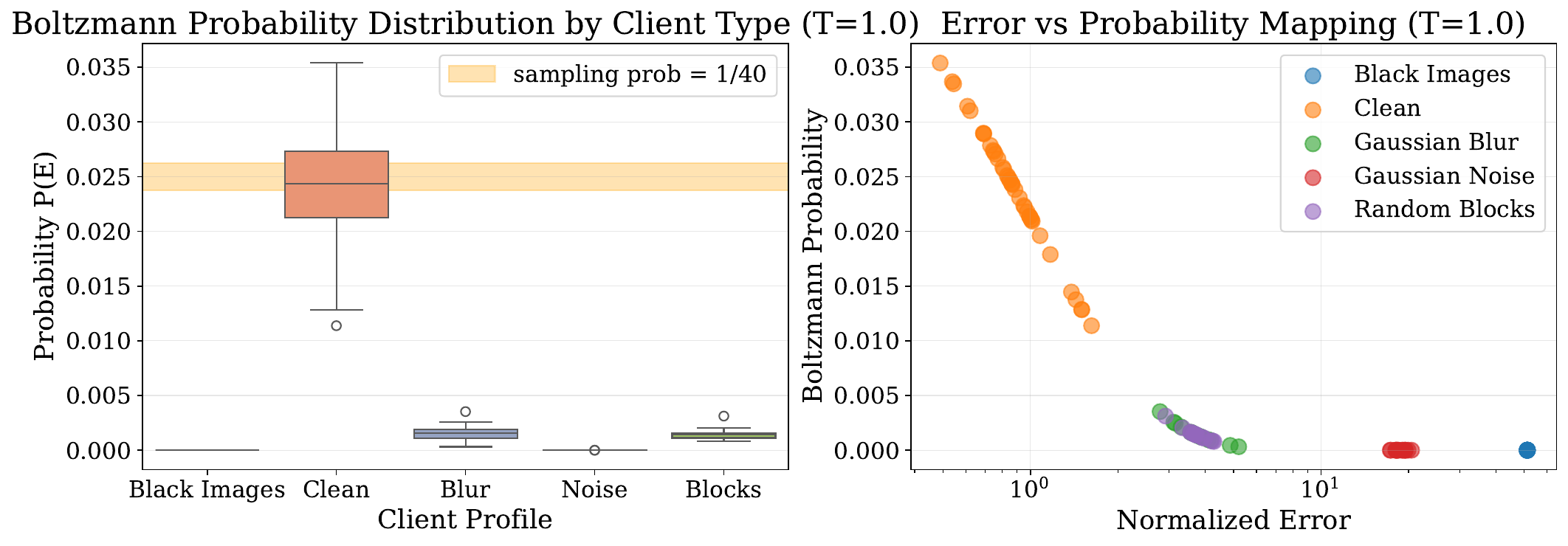}
\caption{\small Boltzmann sampling probabilities $\pi_k$ derived from normalized anomaly
scores on CIFAR-10 under the 60\% corruption scenario, with temperature $\mathcal{T}\!=\!1$.
\textbf{Left:} per-client probabilities grouped by type. \textbf{Right:} $\pi_k$ as a
function of $\mathcal{E}_k$, illustrating the Gibbs mapping from anomaly score to
selection probability. The mean probability assigned to benign clients approaches
$1/40$, which corresponds to the uniform selection probability in the oracle setting
restricted to clean participants, indicating that \shortname{}-Boltzmann recovers
near-oracle sampling behaviour without ground-truth labels. Corrupted clients
exhibiting milder spectral distortions, such as Gaussian blur, receive non-negligible
residual probabilities below $0.5\%$.}
    \label{fig:boltzmann}
\end{figure*}
\begin{table*}[h]
\centering
\caption{\small Global model test accuracy (\%) at communication round 500,
         reported as mean\,{\scriptsize$\pm$\,std} over three independent random seeds.
         Two corruption scenarios are evaluated: \textbf{60\% malicious}
         (40 benign, 15 Gaussian blur, 15 Gaussian noise, 15 Random Block, and 15 Black Images clients) and
         \textbf{90\% malicious} (10 benign, 22 Gaussian blur, 23 Gaussian noise, 22 Random Block, and 23 Black Images clients).
         \textit{FedAvg}~\cite{mcmahan2017communication} selects clients uniformly at random
         from the full pool and serves as the undefended lower bound.
         \textit{Oracle} draws uniformly from ground-truth benign clients only and
         is unavailable in any real deployment; it serves as the performance ceiling
         (note that at 90\% corruption on CIFAR-100, Oracle underperforms FedAvg
         due to the severely limited clean pool of only 10 clients).
         \textit{\shortname{}-Boltzmann} samples clients according to the Gibbs
         distribution $\pi_k$ (Eq.~\eqref{eq:boltzmann}), with temperature $\mathcal{T}$
         controlling the sharpness of exclusion.
         \textit{\shortname{}-Deterministic} admits only clients satisfying
         $\mathcal{E}_k \leq 1+\varepsilon$ (Eq.~\eqref{eq:score}).
         \textbf{Bold}: best result per scenario within each method family.
         \underline{Underline}: Oracle upper bound.}
\label{tab:full_results}
\begin{adjustbox}{max width=\textwidth}
\small
\setlength{\tabcolsep}{4pt}
\begin{tabular}{ll rr rr rr rr rr rr}
\toprule
& & \multicolumn{4}{c}{\textbf{MNIST}}
  & \multicolumn{4}{c}{\textbf{CIFAR-10}}
  & \multicolumn{4}{c}{\textbf{CIFAR-100}} \\
\cmidrule(lr){3-6}\cmidrule(lr){7-10}\cmidrule(lr){11-14}
\textbf{Method} &
  & \multicolumn{2}{c}{60\% mal.}
  & \multicolumn{2}{c}{90\% mal.}
  & \multicolumn{2}{c}{60\% mal.}
  & \multicolumn{2}{c}{90\% mal.}
  & \multicolumn{2}{c}{60\% mal.}
  & \multicolumn{2}{c}{90\% mal.} \\
\cmidrule(lr){3-4}\cmidrule(lr){5-6}
\cmidrule(lr){7-8}\cmidrule(lr){9-10}
\cmidrule(lr){11-12}\cmidrule(lr){13-14}
\midrule

FedAvg~\cite{mcmahan2017communication}
  & {}
  & 96.33 & {\scriptsize$\pm$0.16}
  & 95.10 & {\scriptsize$\pm$0.30}
  & 50.60 & {\scriptsize$\pm$0.77}
  & 45.14 & {\scriptsize$\pm$1.28}
  & 29.97 & {\scriptsize$\pm$0.51}
  & 22.64 & {\scriptsize$\pm$1.94} \\

Oracle
  & {}
  & \ora{97.34} & {\scriptsize$\pm$0.12}
  & \ora{95.29} & {\scriptsize$\pm$0.58}
  & \ora{56.63} & {\scriptsize$\pm$0.34}
  & \ora{46.39} & {\scriptsize$\pm$0.94}
  & \ora{35.82} & {\scriptsize$\pm$0.37}
  & 21.43        & {\scriptsize$\pm$0.53} \\

\midrule
\multicolumn{14}{l}{\textit{\shortname{}-Boltz. (\textbf{ours})}} \\[1pt]

& $\mathcal{T}=0.01$
  & 95.79 & {\scriptsize$\pm$0.14}
  & 95.26 & {\scriptsize$\pm$0.25}
  & 46.60 & {\scriptsize$\pm$0.89}
  & 46.15 & {\scriptsize$\pm$0.54}
  & 22.37 & {\scriptsize$\pm$0.48}
  & 21.62 & {\scriptsize$\pm$0.79} \\

& $\mathcal{T}=0.10$
  & 96.74 & {\scriptsize$\pm$0.40}
  & 95.30 & {\scriptsize$\pm$0.40}
  & 53.09 & {\scriptsize$\pm$1.02}
  & 46.18 & {\scriptsize$\pm$0.38}
  & 30.36 & {\scriptsize$\pm$1.18}
  & 22.00 & {\scriptsize$\pm$0.40} \\

& $\mathcal{T}=0.50$
  & 97.19 & {\scriptsize$\pm$0.25}
  & 95.42 & {\scriptsize$\pm$0.62}
  & 55.72 & {\scriptsize$\pm$0.33}
  & 47.41 & {\scriptsize$\pm$0.99}
  & 35.68 & {\scriptsize$\pm$1.03}
  & 22.52 & {\scriptsize$\pm$1.03} \\

& $\mathcal{T}=1.00$
  & \best{97.41} & {\scriptsize$\pm$0.31}
  & \best{96.45} & {\scriptsize$\pm$0.26}
  & 56.44        & {\scriptsize$\pm$0.63}
  & \best{47.87} & {\scriptsize$\pm$0.24}
  & 34.66        & {\scriptsize$\pm$1.63}
  & \best{24.10} & {\scriptsize$\pm$1.71} \\

& $\mathcal{T}=2.00$
  & 97.37 & {\scriptsize$\pm$0.24}
  & 96.43 & {\scriptsize$\pm$0.28}
  & 54.75 & {\scriptsize$\pm$0.61}
  & 47.47 & {\scriptsize$\pm$0.37}
  & 34.92 & {\scriptsize$\pm$0.49}
  & 22.61 & {\scriptsize$\pm$4.47} \\

& $\mathcal{T}=5.00$
  & 97.38 & {\scriptsize$\pm$0.11}
  & 95.51 & {\scriptsize$\pm$0.34}
  & 52.82 & {\scriptsize$\pm$1.99}
  & 46.32 & {\scriptsize$\pm$0.94}
  & 32.30 & {\scriptsize$\pm$0.46}
  & 24.62 & {\scriptsize$\pm$0.67} \\

\midrule
\multicolumn{14}{l}{\textit{\shortname{}-Det. (\textbf{ours})}} \\[1pt]

& $\varepsilon=0.05$
  & 96.14 & {\scriptsize$\pm$1.98}
  & 94.42 & {\scriptsize$\pm$0.90}
  & 53.56 & {\scriptsize$\pm$1.99}
  & 43.17 & {\scriptsize$\pm$2.33}
  & 31.28 & {\scriptsize$\pm$2.44}
  & 18.03 & {\scriptsize$\pm$1.72} \\

& $\varepsilon=0.10$
  & 96.66 & {\scriptsize$\pm$0.74}
  & 94.42 & {\scriptsize$\pm$1.17}
  & 54.90 & {\scriptsize$\pm$0.08}
  & 44.16 & {\scriptsize$\pm$2.21}
  & 31.94 & {\scriptsize$\pm$2.47}
  & 18.95 & {\scriptsize$\pm$0.76} \\

& $\varepsilon=0.30$
  & 97.19 & {\scriptsize$\pm$0.21}
  & 95.06 & {\scriptsize$\pm$0.64}
  & 55.87 & {\scriptsize$\pm$1.07}
  & 44.49 & {\scriptsize$\pm$0.50}
  & 32.44 & {\scriptsize$\pm$2.63}
  & 20.78 & {\scriptsize$\pm$0.91} \\

& $\varepsilon=0.50$
  & 97.10 & {\scriptsize$\pm$0.35}
  & 95.47 & {\scriptsize$\pm$0.14}
  & 56.80 & {\scriptsize$\pm$1.12}
  & 45.55 & {\scriptsize$\pm$1.20}
  & 34.54 & {\scriptsize$\pm$1.27}
  & 21.38 & {\scriptsize$\pm$0.38} \\

& $\varepsilon=1.00$
  & 97.22 & {\scriptsize$\pm$0.18}
  & 95.35 & {\scriptsize$\pm$0.04}
  & \best{57.07} & {\scriptsize$\pm$0.97}
  & 46.04        & {\scriptsize$\pm$0.14}
  & 34.75 & {\scriptsize$\pm$0.42}
  & 21.32        & {\scriptsize$\pm$0.88} \\

& $\varepsilon=2.00$
  & {97.39} & {\scriptsize$\pm$0.08}
  & 95.26        & {\scriptsize$\pm$0.16}
  & 55.71        & {\scriptsize$\pm$1.12}
  & 46.00        & {\scriptsize$\pm$0.66}
  & \best{35.48}        & {\scriptsize$\pm$1.66}
  & {21.57} & {\scriptsize$\pm$0.20} \\

\bottomrule
\end{tabular}
\end{adjustbox}
\end{table*}

\begin{table*}[t]
\centering
\caption{\small Client detection performance (F1, precision, recall, accuracy in \%),
         mean\,{\scriptsize$\pm$\,std} over three random seeds, under 40\% and 90\%
         malicious client scenarios. Corruptions consist of equal proportions of
         Gaussian blur and Gaussian noise.
         \textsc{Waffle}-FT and \textsc{Waffle}-WST~\cite{licciardi2025waffle} rely on a
         supervised classifier trained with labelled attack profiles;
         \shortname{}-Boltzmann and \shortname{}-Deterministic are fully unsupervised,
         requiring no prior knowledge of the attack type.
         Best hyperparameters $\mathcal{T}^*$ and $\varepsilon^*$ are selected per scenario by
         mean F1 across seeds.
}
\label{tab:detection_comparison}
\begin{adjustbox}{max width=\textwidth}
\small
\setlength{\tabcolsep}{3.5pt}
\begin{tabular}{cl
                rrrr
                rrrr}
\toprule
& & \multicolumn{4}{c}{\textbf{CIFAR-10}}
  & \multicolumn{4}{c}{\textbf{CIFAR-100}} \\
\cmidrule(lr){3-6}\cmidrule(lr){7-10}
\textbf{} & \textbf{Method}
  & F1 & Prec. & Rec. & Acc.
  & F1 & Prec. & Rec. & Acc. \\
\midrule

\multirow{4}{*}{\rotatebox[origin=c]{90}{\small 40\% mal.}}
& \textsc{Waffle}-FT~\cite{licciardi2025waffle}
  & 80.2{\scriptsize$\pm$2.6} & 69.1{\scriptsize$\pm$2.6} & 96.1{\scriptsize$\pm$2.6} & 67.0{\scriptsize$\pm$2.6}
  & 55.1{\scriptsize$\pm$3.2} & 40.5{\scriptsize$\pm$2.6} & {89.7}{\scriptsize$\pm$2.6} & 44.1{\scriptsize$\pm$2.6} \\

& \textsc{Waffle}-WST~\cite{licciardi2025waffle}
  & 95.2{\scriptsize$\pm$1.0} & 97.6{\scriptsize$\pm$1.0} & 92.9{\scriptsize$\pm$1.0} & 96.1{\scriptsize$\pm$1.0}
  & 83.0{\scriptsize$\pm$1.2} & 93.1{\scriptsize$\pm$1.2} & 75.1{\scriptsize$\pm$1.2} & 87.0{\scriptsize$\pm$1.2} \\

& \shortname{}-Boltz. (\textbf{ours}) ($\mathcal{T}^*\!=\!0.5$)
  & \best{98.8}{\scriptsize$\pm$0.0} & 97.6{\scriptsize$\pm$0.0} & \best{100.0}{\scriptsize$\pm$0.0} & \best{99.0}{\scriptsize$\pm$0.0}
  & \best{98.8}{\scriptsize$\pm$1.2} & \best{97.6}{\scriptsize$\pm$2.4} & \best{100.0}{\scriptsize$\pm$0.0} & \best{99.0}{\scriptsize$\pm$1.0} \\

& \shortname{}-Det. (\textbf{ours}) ($\varepsilon^*\!=\!1.0$)
  & 98.0{\scriptsize$\pm$0.7} & \best{97.8}{\scriptsize$\pm$0.9} & \best{100.0}{\scriptsize$\pm$0.0} & 98.3{\scriptsize$\pm$0.6}
  & 98.0{\scriptsize$\pm$1.4} & 96.0{\scriptsize$\pm$2.6} & \best{100.0}{\scriptsize$\pm$0.0} & 98.3{\scriptsize$\pm$1.2} \\

\midrule

\multirow{4}{*}{\rotatebox[origin=c]{90}{\small 90\% mal.}}
& \textsc{Waffle}-FT~\cite{licciardi2025waffle}
  & 93.3{\scriptsize$\pm$1.6} & 89.2{\scriptsize$\pm$1.6} & {95.7}{\scriptsize$\pm$1.6} & 86.2{\scriptsize$\pm$1.6}
  & 89.0{\scriptsize$\pm$1.6} & 88.2{\scriptsize$\pm$1.6} &{88.4}{\scriptsize$\pm$1.6} & 81.1{\scriptsize$\pm$1.6} \\

& \textsc{Waffle}-WST~\cite{licciardi2025waffle}
  & 91.1{\scriptsize$\pm$0.5} & \best{100.0}{\scriptsize$\pm$0.0} & 83.8{\scriptsize$\pm$0.5} & 87.0{\scriptsize$\pm$0.5}
  & {88.1}{\scriptsize$\pm$0.3} & \best{100.0}{\scriptsize$\pm$0.0} & 68.3{\scriptsize$\pm$0.3} & 72.2{\scriptsize$\pm$0.3} \\

& \shortname{}-Boltz. (\textbf{ours}) ($\mathcal{T}^*\!=\!0.1$)
  & {99.3}{\scriptsize$\pm$0.8} & 98.6{\scriptsize$\pm$1.6} & \best{100.0}{\scriptsize$\pm$0.0} & {98.7}{\scriptsize$\pm$1.5}
  & 99.3{\scriptsize$\pm$0.6} & 98.6{\scriptsize$\pm$1.3} & \best{100.0}{\scriptsize$\pm$0.0} & {98.7}{\scriptsize$\pm$1.2} \\

& \shortname{}-Det. (\textbf{ours}) ($\varepsilon^*\!=\!1.0$)
  & \best{100.0}{\scriptsize$\pm$0.0} & \best{100.0}{\scriptsize$\pm$0.0} & \best{100.0}{\scriptsize$\pm$0.0} & \best{100.0}{\scriptsize$\pm$0.0}
  & \best{99.6}{\scriptsize$\pm$0.3} & 99.3{\scriptsize$\pm$0.6} & \best{100.0}{\scriptsize$\pm$0.0} & \best{99.3}{\scriptsize$\pm$0.6} \\

\bottomrule
\end{tabular}
\end{adjustbox}
\end{table*}


\subsection{Experimental Setup}
\label{sec:setup}

\paragraph*{Datasets and Partitioning}
\label{sec:datasets}

Experiments are conducted on three standard image classification benchmarks of increasing difficulty \cite{caldas2018leaf}. \textbf{MNIST}~\cite{lecun1998mnist} provides 60{,}000 grayscale training images of handwritten digits ($28\!\times\!28$ pixels, 10 classes), normalized with $\mu\!=\!0.1307$ and $\sigma\!=\!0.3081$. \textbf{CIFAR-10}~\cite{cifar10} consists of 50{,}000 RGB training images ($32\!\times\!32$ pixels) across 10 object categories, with per-channel normalization $\boldsymbol{\mu}\!=\!(0.4914,\,0.4822,\,0.4465)$ and $\boldsymbol{\sigma}\!=\!(0.2470,\,0.2435,\,0.2616)$. \textbf{CIFAR-100}~\cite{krizhevsky2009learning} shares the same resolution and training set size but spans 100 fine-grained classes, constituting a substantially harder recognition problem. For all three datasets, $20\%$ of the training set is withheld as a server-side auxiliary set $D_{\text{aux}}$, used exclusively for offline autoencoder training as described in Section~\ref{subsec:dae}. The remaining $80\%$ constitutes the client pool from which local datasets are drawn.
\paragraph*{Federated Scenario and Corruption Model}
\label{sec:partition}

We instantiate a cross-device FL scenario~\cite{mcmahan2017communication} with $K\!=\!100$ clients, each assigned a local dataset of $n_k\!=\!200$ images sampled uniformly without replacement from the client pool. This non-overlapping assignment yields a controlled yet realistic simulation of edge devices. To stress-test \shortname{} across qualitatively different threat regimes, we design two experimental scenarios that differ in the proportion of corrupted clients. In the \textbf{moderate-corruption scenario}, $60\%$ of clients are corrupted, placing the federation just above the theoretical failure threshold of Byzantine-robust aggregation methods~\cite{blanchard2017machine}. In the \textbf{severe-corruption scenario}, the fraction of corrupted clients is raised to $90\%$, leaving only 10 benign participants and constituting an extreme adversarial pressure under which virtually all existing defenses collapse~\cite{guerraoui2018hidden, licciardi2025waffle}. In both scenarios, corrupted clients are partitioned equally across four corruption types. Clients subject to \emph{Gaussian noise} receive additive white Gaussian perturbations with $\sigma\!=\!2.0$, modelling sensor degradation or low signal-to-noise acquisition~\cite{fang2020local}. Clients subject to \emph{Gaussian blur} have their images convolved with a Gaussian kernel (kernel size $\!=\!19$, variance $\!=\!20$), simulating focus degradation or motion artefacts. Clients subject to \emph{random pixel masking} have $70\%$ of their pixels zeroed uniformly at random, representing packet loss or partial sensor occlusion. Finally, clients transmitting \emph{black images} have all pixel values set to zero, modelling completely unresponsive or dead devices. Each corruption type is designed to produce a distinct distortion, enabling the \shortname{} autoencoder to assign anomalous reconstruction scores $\mathcal{E}_k > 1 + \epsilon$ to corrupted clients without any prior knowledge of the corruption nature, where $\epsilon >0$ is a tolerance parameter.

\paragraph*{Model Architectures and Training Configuration}
\label{sec:architectures}

The global model is chosen to match the representational demands of each dataset. For MNIST, a lightweight convolutional network with two convolutional layers (16 and 32 feature maps), max-pooling, and a 128-unit fully connected classifier suffices. For CIFAR-10, the same architecture is widened to 32 and 64 feature maps respectively. For CIFAR-100, the increased class granularity requires greater capacity; we adopt ResNet-9~\cite{he2016deep}, a compact residual architecture with two skip-connected blocks and a 512-unit classification head, providing sufficient expressivity while remaining tractable in a federated setting with limited per-client data. At each communication round, $C\!=\!10$ clients participate (10\% participation rate). Each selected client performs $E\!=\!1$ local epoch of mini-batch SGD with learning rate $\eta\!=\!0.01$ and batch size 32. The server aggregates updates via standard FedAvg weight averaging~\cite{mcmahan2017communication}. The \shortname{} autoencoder is trained for 100 epochs using the Adam optimiser~\cite{kingma2014adam} with $\eta\!=\!10^{-3}$, bottleneck dimension $d_z\!=\!16$, and denoising noise level $\sigma_{\text{noise}}\!=\!0.1$. The WST is computed via the Kymatio library~\cite{andreux2020kymatio} with $J\!=\!3$, $L\!=\!8$ for $32\!\times\!32$ images and $J\!=\!4$, $L\!=\!8$ for $28\!\times\!28$ images, retaining only first-order coefficients, as motivated in \cite{licciardi2025waffle}. All experiments are run for 500 communication rounds with three fixed random seeds, controlling dataset splits, client assignment, and per-round sampling. 

\subsection{Results and Discussion}\label{sec:results}

\paragraph*{Spectral Representation: \shortname{}-WST vs.\ FT}

We first justify the architectural choice of the Wavelet Scattering Transform as the
feature extractor backbone.
Figure~\ref{fig:wstvsft} contrasts the normalised anomaly scores $\mathcal{E}_k$
produced by the proposed WST backbone against those yielded by a standard Fourier
Transform (FT) backbone, configured with optimal parameters derived from~\cite{licciardi2025waffle}, on CIFAR-10 under the 60\% corruption scenario.
Under the FT representation, anomaly scores for clients subjected to Gaussian blur and
random pixel masking overlap substantially with the benign client distribution on both
sides of the detection threshold $\tau = 1$, rendering these two structurally subtle
corruption types effectively indistinguishable from clean clients.
This failure is consistent with the well-established theoretical instability of the
Fourier modulus to local spatial deformations: a small diffeomorphic perturbation that
leaves pixel-space amplitude bounded can induce arbitrarily large high-frequency
spectral displacement, preventing the denoising autoencoder from learning a stable and
discriminative reconstruction manifold over FT-encoded inputs~\cite{mallat2012group}.
In contrast, the WST backbone yields a strict and geometrically stable separation: the
anomaly scores of all 40 clean clients remain tightly concentrated below $\tau = 1$
with low inter-client variance, while all four corruption profiles—including the
structurally subtle Gaussian blur and random pixel masking—are mapped consistently and
with large margin above the threshold.

\paragraph*{Boltzmann Sampling Behaviour}
Figure~\ref{fig:boltzmann} displays that, at $\mathcal{T}\!=\!1$, the mean selection
probability of benign clients is approximately $1/40$—the oracle uniform rate over the
clean pool—without access to ground-truth labels. High-anomaly clients are
exponentially suppressed via $\pi_k \propto \exp(-\mathcal{E}_k/\mathcal{T})$, while
mildly distorted clients retain residual probabilities below $0.5\%$, preserving
marginal gradient diversity when the benign pool is small.

\paragraph*{Global Model Classification Performance}
Table~\ref{tab:full_results} reports test accuracy at round 500. FedAvg degrades
modestly on MNIST ($96.33\%\!\to\!95.10\%$) but substantially on CIFAR-100
($29.97\%\!\to\!22.64\%$) as the malicious fraction rises from 60\% to 90\%,
reflecting increasing sensitivity to corrupted gradients with task complexity. The
Oracle underperforms FedAvg on CIFAR-100 at 90\% corruption ($21.43\%$ vs.\
$22.64\%$), since only 10 benign clients yield an expected cohort of one clean
participant per round—too sparse for effective 100-class training.
\shortname{}-Boltzmann ($\mathcal{T}\!=\!1.0$) achieves best results across most
settings ($97.41\%$ MNIST, $56.44\%$ CIFAR-10 at 60\%), and uniquely surpasses both
Oracle and FedAvg on CIFAR-100 at 90\% ($24.10\%$) via soft sampling under benign
scarcity. \shortname{}-Deterministic ($\varepsilon\!=\!1.0$) matches or exceeds the
Oracle on CIFAR-10 ($57.07\%$) and CIFAR-100 ($35.48\%$ at $\varepsilon\!=\!2.0$)
under 60\% corruption, but exhibits higher variance in the severe regime, making it
more sensitive to hyperparameter choice than the Boltzmann variant.

\paragraph*{Client Detection: \shortname{} vs.\ \textsc{Waffle}}
Table~\ref{tab:detection_comparison} evaluates binary detection against the supervised
\textsc{Waffle} baselines~\cite{licciardi2025waffle}.
Operating fully unsupervised, \shortname{} outperforms both \textsc{Waffle}-FT and
\textsc{Waffle}-WST on every metric. At 40\% corruption, \shortname{}-Boltzmann
($\mathcal{T}^*\!=\!0.5$) achieves $98.8\%$ F1 and $100\%$ recall on both CIFAR-10
and CIFAR-100, versus $95.2\%$ and $83.0\%$ for \textsc{Waffle}-WST. Under 90\%
corruption, \textsc{Waffle}-WST suffers a recall collapse to $68.3\%$ on CIFAR-100
under the imbalanced client distribution, while \shortname{}-Deterministic
($\varepsilon^*\!=\!1.0$) achieves perfect $100\%$ F1 on CIFAR-10 and $99.6\%$ on
CIFAR-100—without any prior knowledge of the attack type.
\vspace{-1.em}
\section{Conclusion}
We present \shortname{}, an offline detection algorithm for
FL that grounds client screening in the geometry
of a learned benign WST manifold rather than in labeled attack
profiles. This design choice removes the supervised assumption
entirely, enabling generalization to unseen and composite
corruptions without retraining, even when benign clients are a
minority. The Boltzmann sampling strategy further decouples
defensive filtering from hard exclusion, gracefully sustaining
convergence when the clean client pool is critically small.
Together, these contributions advance offline FL defense from a
supervised, threat-specific paradigm to a fully unsupervised,
attack-agnostic one.
\vspace{-1em}
\section*{Acknowledgment}
A.L.\ is a member of GNFM-INdAM and was supported by the PNRR-NGEU project of the Italian MUR (DM~117/2023), and thanks Prof.\ Lamberto Rondoni for valuable discussions. 
\bibliography{reference}
\bibliographystyle{ieeetr}
\end{document}